\documentclass{article}

\usepackage[main, final]{neurips_2026}

\usepackage[utf8]{inputenc} 
\usepackage[T1]{fontenc}    
\usepackage{hyperref}       
\usepackage{url}            
\usepackage{booktabs}       
\usepackage{amsfonts}       
\usepackage{nicefrac}       
\usepackage{microtype}      
\usepackage{xcolor}         

\definecolor{darkblue}{rgb}{0, 0, 0.5}
\hypersetup{
    colorlinks=true,
    citecolor=darkblue,
    linkcolor=darkblue,
    urlcolor=darkblue
}
\hypersetup{
    colorlinks=true,
    citecolor=blue,
    linkcolor=blue,
    urlcolor=blue
}

\usepackage{enumitem}       
\usepackage{subfig}         
\usepackage{multirow}       %
\usepackage{xspace}         
\usepackage{wrapfig}        
\usepackage{placeins}       
\usepackage{makecell}       
\usepackage{graphicx}       
\usepackage{colortbl}
\usepackage{bbm}            
\usepackage{adjustbox}

\usepackage{bm}
\usepackage{pifont}         
\usepackage{bbding}         
\usepackage{xurl}           
\NewDocumentCommand{\drsh}{ O{0.6em} O{0.5em} O{0.65pt} O{rounded corners=1.6pt} }{%
  \mathrel{%
  \hspace{0.3em}
    \tikz[baseline=-0.6em]{%
      \draw[->, line width=#3, #4] (0,0) -- (0,-#2) -- (#1,-#2);
    }%
  }%
}

\usepackage{amsmath}
\usepackage{amssymb}
\usepackage{mathtools}
\usepackage{amsthm}

\theoremstyle{plain}

\theoremstyle{definition}

\theoremstyle{remark}

\usepackage[textsize=tiny]{todonotes}

\usepackage{algorithm}
\usepackage{algorithmicx}
\usepackage{algpseudocode}

\usepackage{array}
\newcolumntype{N}{c@{\hspace{2pt}}}    
\newcolumntype{L}[1]{>{\raggedright\arraybackslash}p{#1}}
\newcolumntype{C}[1]{>{\centering\arraybackslash}p{#1}}

\setlist[itemize]{leftmargin=20pt}
\setlist[enumerate]{leftmargin=20pt}

\usepackage[most,skins,theorems,breakable]{tcolorbox}
\tcbset{
  mybox/.style={
    width=\linewidth,
    top=14pt,
    bottom=7pt,
    colback=cyan!5!white,
    colframe=black,
    colbacktitle=black,
    enhanced,
    center,
    breakable,
    attach boxed title to top left={yshift=-0.13in,xshift=0.15in},
    boxed title style={boxrule=0pt,colframe=white,},
  }
}
\newtcolorbox{mybox}[2][]{mybox,title=#2,#1}

\newcommand{\ourmethod}{Archer\xspace}

\newcommand{\mycolor}{cyan!10}

\newcommand{\DSOneB}{DeepSeek-R1-Distill-Qwen-1.5B\xspace}
\newcommand{\QFourB}{Qwen3-4B\xspace}
\newcommand{\QEightB}{Qwen3-8B\xspace}
\newcommand{\QThirtyB}{Qwen3-30B-A3B\xspace}

\usepackage{circledsteps}

\newcommand\blfootnote[1]{%
  \begingroup
  \renewcommand\thefootnote{}\footnote{#1}%
  \addtocounter{footnote}{-1}%
  \endgroup
}

\title{Stabilizing Knowledge, Promoting Reasoning: Dual-Token Constraints for RLVR}

\author{%
\textbf{Runze Liu$^{1*}$} \quad
\textbf{Jiakang Wang$^{2*}$} \quad
\textbf{Fuzheng Zhang$^{2}$} \quad
\textbf{Xiu Li$^{3}$} \quad \\
\textbf{Guorui Zhou$^{2}$} \quad
\textbf{Kun Gai$^{2}$} \quad 
\textbf{Ling Pan$^{1\dag}$}\\[1.2ex]
$^{1}$Hong Kong University of Science and Technology,
$^{2}$Kuaishou Technology,
$^{3}$Tsinghua University
}

\begin{document}

\blfootnote{$^*$ Equal contribution}
\blfootnote{$^\dag$ Corresponding author: Ling Pan (lingpan@ust.hk)}

\maketitle

\begin{abstract}
Reinforcement Learning with Verifiable Rewards (RLVR) has become an effective post-training method for improving the reasoning abilities of Large Language Models (LLMs). However, existing methods mainly apply uniform optimization constraints across all tokens, ignoring their heterogeneous roles. Prior work shows that high-entropy tokens are closely tied to reasoning, while low-entropy tokens primarily encode factual knowledge, and recent approaches attempt to exploit this distinction by isolating token updates via masking or asynchronous training. We argue that retaining training signals across token types can lead to more effective learning. To address this, we propose \textbf{Archer}, an entropy-aware RLVR framework with \textbf{dual-token constraints} that preserves joint optimization while modulating update strength across token types. Our method introduces response-level entropy normalization for stable token classification and applies differentiated clipping ranges and KL regularization to encourage exploration on reasoning tokens while preserving knowledge tokens. Experiments on mathematical reasoning and code generation benchmarks show that Archer consistently outperforms strong baselines across multiple model scales, improving both \textit{pass@1} and \textit{pass@K} performance. These results highlight the importance of respecting sequence-level dependencies when designing fine-grained RL optimization strategies for LLMs. The code is available at~\url{https://github.com/wizard-III/ArcherCodeR}.
\end{abstract}

\section{Introduction}
\label{sec:introduction}

Large Language Models~(LLMs) have shown strong capabilities across a wide range of domains, demonstrated by models like OpenAI's ``o'' series~\citep{o1, o3} and DeepSeek-R1~\citep{DeepSeek-R1}. While pre-training enables LLMs to obtain vast amount of world knowledge, post-training techniques such as Reinforcement Learning~(RL)~\citep{DeepSeek-R1, Kimi-k1.5, Qwen3, zhang2025survey} and test-time scaling~\citep{snell2025scaling, liu2025can} are crucial for enhancing their reasoning abilities. Compared to approaches like Monte Carlo Tree Search~(MCTS)~\citep{TS-LLM} and Process Reward Modeling~\citep{PRM800K, Math-Shepherd, GenPRM}, Reinforcement Learning with Verifiable Rewards (RLVR) has emerged as a simple yet effective way to further improve the reasoning abilities of LLMs~\citep{GRPO, DAPO, MARTI}.

Recent studies have revealed that RL mainly improves reasoning by better integrating and organizing the model's existing abilities, such as reflection and planning, rather than directly changing the model's factual memory or basic skills (e.g., arithmetic)~\citep{gandhi2025cognitive, vassoyan2025ignore, li2025llms}. In particular, improvements are largely concentrated on tokens that mediate logical transitions (e.g., decision points and reasoning connectors) rather than tokens encoding factual knowledge. Prior analyses have shown that these reasoning-critical tokens tend to exhibit higher entropy, while knowledge-related tokens are typically low-entropy and more stable~\citep{80/20, cheng2025reasoning, Rejuvenation}.
This observation raises a fundamental question:

\textit{How should RL algorithms account for the heterogeneous functional roles of tokens during optimization?}

Existing RLVR methods largely treat all tokens uniformly, applying identical update constraints across the entire response~\citep{GRPO, DAPO}. To address this limitation, several recent works propose to explicitly differentiate token types based on entropy or related statistics, for example by masking gradients for low-entropy tokens or updating different token groups asynchronously~\citep{80/20, Entropy-Mechanism, Lopti}. While these approaches acknowledge token heterogeneity, they rely on \emph{isolating} subsets of tokens during training.
However, we argue that such isolation is fundamentally misaligned with the sequential nature of language modeling. In LLMs, tokens are generated autoregressively, and the optimization signal for a given token depends on its surrounding context. As a result, token-level updates are inherently \textbf{coupled across the sequence}. Completely masking certain tokens removes their direct training signals and, in our controlled ablation, leads to lower performance (Section~\ref{subsubsec:ablation_masking}).

To address these limitations, we introduce \textbf{Archer}, which adaptively constructs a token-conditioned trust region geometry while preserving joint optimization. Our key insight is that token heterogeneity should shape the permissible policy movement at each position, rather than by masking tokens or optimizing tokens uniformly.
Locally, within each policy improvement step, trust regions should be enlarged for uncertain, reasoning-sensitive tokens, allowing the policy to flexibly explore alternative transitions, while being contracted for stable, knowledge-intensive positions to prevent large, destructive policy updates to reliable predictions. Globally, across training, reasoning-sensitive tokens should be allowed to drift farther from the reference policy to acquire reward-aligned reasoning behaviors, whereas knowledge-stable tokens should remain more tightly anchored to the pretrained distribution to preserve factual priors at the global level. Archer realizes these principles through a simple yet effective dual-anchor trust-region design.
Instead of introducing auxiliary networks with additional computational overhead, we infer token roles by utilizing intra-response entropy quantiles zero-shot. Based on this information-theoretic proxy, we adaptively reshape the trust region along two complementary dimensions.
The local principle is instantiated as a token-conditioned proximal constraint, which controls the permissible update magnitude at each position within a single optimization step through token-conditioned clipping. The global principle is instantiated as a token-conditioned reference-policy anchor, which controls the long-horizon drift of each position from the base model through token-conditioned KL regularization. Together, this dual-anchor mechanism effectively trades off between flexible reasoning and rigid factual preservation. 

In summary, our contributions are as follows:
\begin{itemize}
\item We identify the limitations of token isolation strategies and entropy statistics in RLVR and introduce an intra-response entropy quantile scheme that infers token roles zero-shot.
\item We propose Archer, a token-conditioned dual-anchor trust-region framework that preserves joint autoregressive optimization while adaptively allocating policy movement budgets across tokens. Archer couples a local proximal anchor for step-wise update control with a global reference-policy anchor for long-horizon drift control.
\item We demonstrate consistent improvements over strong baselines on both mathematical reasoning and code generation tasks with several model sizes.
\end{itemize}

\section{Related Work}

\subsection{Reinforcement Learning for Large Language Models}

Previous works have shown that RL, particularly Reinforcement Learning from Human Feedback (RLHF)~\citep{christiano2017deep, MRN}, is an effective tool for aligning LLMs with human preferences~\citep{ouyang2022training, bai2022training}. With the recent success of scaling RL in LLMs~\citep{o1, DeepSeek-R1, Kimi-k1.5}, RLVR has emerged as an effective method to improve the reasoning ability of LLMs using rule-based rewards. However, approaches like GRPO~\citep{GRPO} and its extensions~\citep{DAPO, Dr-GRPO, GPG, VAPO, Skywork-OR1, ASPO} rely on response-level learning signals, which uniformly assign the same advantage value to all tokens within a response. This uniform treatment overlooks the distinct roles tokens play during reasoning (e.g., factual recall vs. logical inference), potentially leading to suboptimal learning at critical reasoning steps and limiting overall performance gains. Although process-based RL~\citep{VinePPO, PRIME, RL-Tango} and unsupervised RL~\citep{EM-RL, cheng2025reasoning} provide fine-grained rewards for RL optimization, they still lack consideration for the functions of different tokens.

\subsection{Critical Token Analysis in RL for Reasoning}

Several recent studies have provided token-level analyses of RLVR training~\citep{Lopti, Entropy-Mechanism, 80/20, cheng2025reasoning}.
\citet{Lopti} observe that low-probability tokens, often exhibiting high entropy, dominate the RL updates and the update of high-probability tokens are suppressed.
\citet{Entropy-Mechanism} show that changes in policy entropy are linked to the covariance between action probabilities and advantages.
\citet{80/20} identify high-entropy tokens, referred to as ``forking tokens'', as logical connectors.
\citet{cheng2025reasoning} further associate high-entropy tokens with reasoning-related behaviors, such as logical transitions and self-reflection.
Unlike prior works that either completely isolate low-entropy tokens~\citep{80/20} or high-covariance tokens~\citep{80/20, Entropy-Mechanism}, or train them separately~\citep{Lopti}, our approach employs joint training. While we similarly utilize entropy to distinguish between logic-oriented and knowledge-oriented tokens, we avoid direct filtering or separation. Instead, we apply differentiated training constraints, enabling us to preserve the capabilities of the base model while simultaneously encouraging more effective exploration during training.

\section{Preliminaries}
\label{sec:preliminaries}

Group Relative Policy Optimization~(GRPO)~\citep{GRPO} proposes an alternative to the value-based advantage estimation used in Proximal Policy Optimization~(PPO)~\citep{PPO}. Instead of learning a value model, GRPO estimates advantages by sampling multiple rollouts per prompt. Specifically, for a given prompt $q$, GRPO generates a group of responses $\{o^1, o^2, \ldots, o^G\}$ and computes the corresponding rewards $\{R^1, R^2, \ldots, R^G\}$. The advantage is then calculated as $\hat{A}_t^i = \frac{R^i - \operatorname{mean}(\{R^i\}_{i=1}^G)}{\operatorname{std}(\{R^i\}_{i=1}^G)}$. The GRPO loss is computed as:
\begin{equation}
\begin{aligned}
    \mathcal{J}_{\text{GRPO}}(\theta) = &\ \mathbb{E}_{q \sim \mathcal{D}, \{o^i\}_{i=1}^G \sim \pi_{\theta_{\text{old}}}(\cdot \mid q)} \\
    & \left[ \frac{1}{G} \sum_{i=1}^G \frac{1}{|o^i|} \sum_{t=1}^{|o^i|} \bigg( \min\left( r_t^i(\theta) \hat{A}_t^i, \operatorname{clip}\big( r_t^i(\theta), 1-\varepsilon, 1+\varepsilon \big) \hat{A}_t^i \right) - \beta \mathbb{D}_{\text{KL}}(\pi_\theta \| \pi_{\text{ref}}) \bigg) \right],
\end{aligned}
\end{equation}
where $r_t^i = \frac{\pi_{\theta}(o_t^i \mid q, o_{<t}^i)}{\pi_{\theta_{\text{old}}}(o_t^i \mid q, o_{<t}^i)}$ denotes the importance sampling ratio, and $\beta$ is a coefficient weighting the Kullback–Leibler~(KL) divergence between the current policy $\pi_\theta$ and the reference policy $\pi_{\text{ref}}$.

\section{Method}
\label{sec:method}

In this section, we introduce \ourmethod, a novel RLVR approach with entropy-aware dual-token constraints. We begin by describing entropy-based method for identifying critical tokens~(Section~\ref{subsec:token_identification}). Next, we discuss the limitations of prior methods in handling low-entropy tokens and motivate our approach for response-level entropy statistics~(Section~\ref{subsubsec:response_level_entropy}). Finally, we detail how \ourmethod improves upon core constraints (clipping and KL) in previous RL algorithms by disentangling token-level optimization~(Section~\ref{subsubsec:our_method}).

\begin{figure*}[!ht]
\centering
\begin{tabular}{cc}
\subfloat[\centering High-Entropy Tokens]{\centering\includegraphics[width=0.46\linewidth]{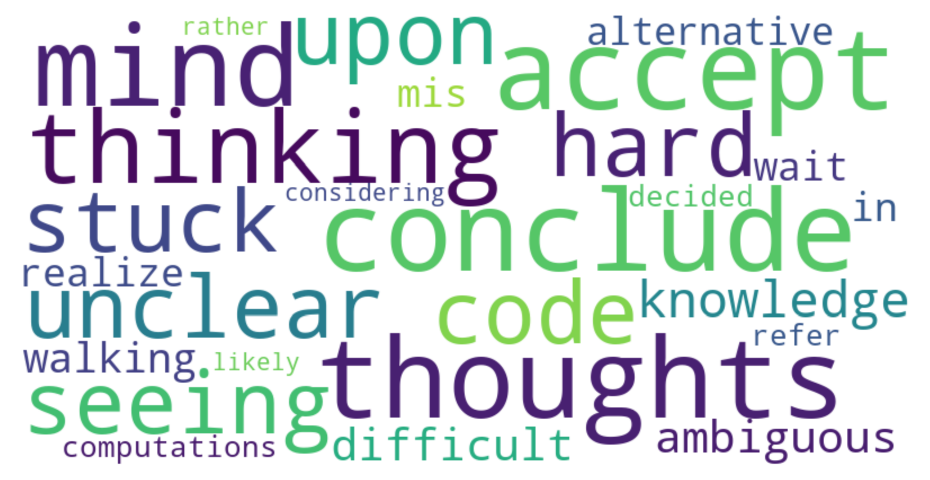}}\label{fig:wordcloud_high_entropy}
\hspace{-0.8em}
& \subfloat[\centering Low-Entropy Tokens]{\includegraphics[width=0.46\linewidth]{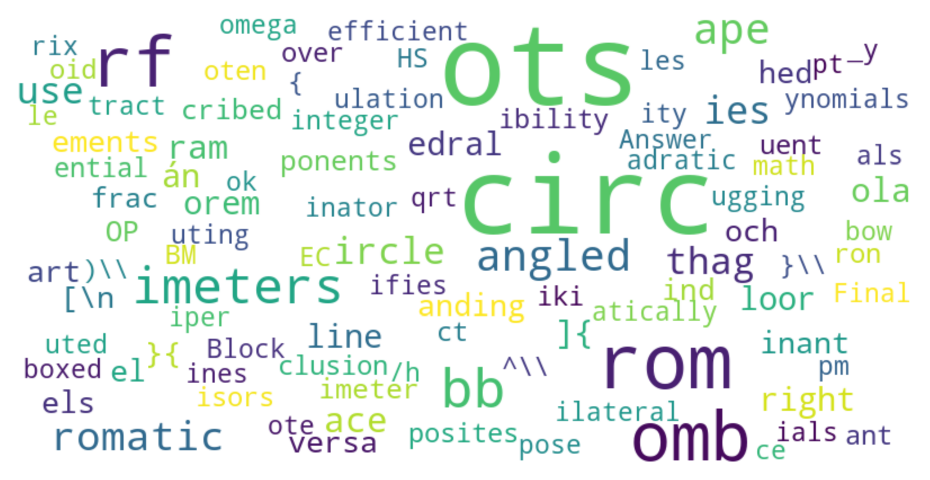}}\label{fig:wordcloud_low_entropy}
\end{tabular}
\caption{Word cloud visualization of a batch of responses: (a) High-entropy tokens; (b) Low-entropy tokens.}
\label{fig:wordcloud}
\end{figure*}

\begin{figure*}[!ht]
\centering
\begin{tabular}{cc}
\hspace{-1.1em}
\subfloat[\centering Prompt-level vs. Batch-level]{\centering\includegraphics[width=0.45\linewidth]{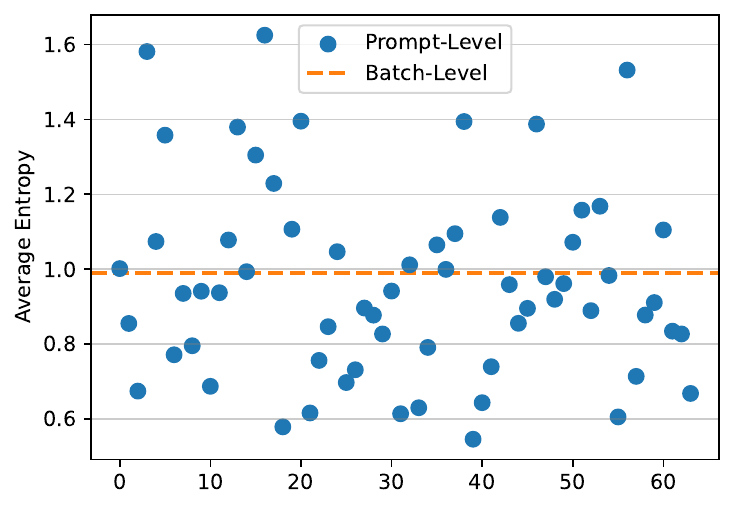}}\label{fig:prompt_avg_entropy}
& \subfloat[\centering Response-level vs. Prompt-level]{\includegraphics[width=0.45\linewidth]{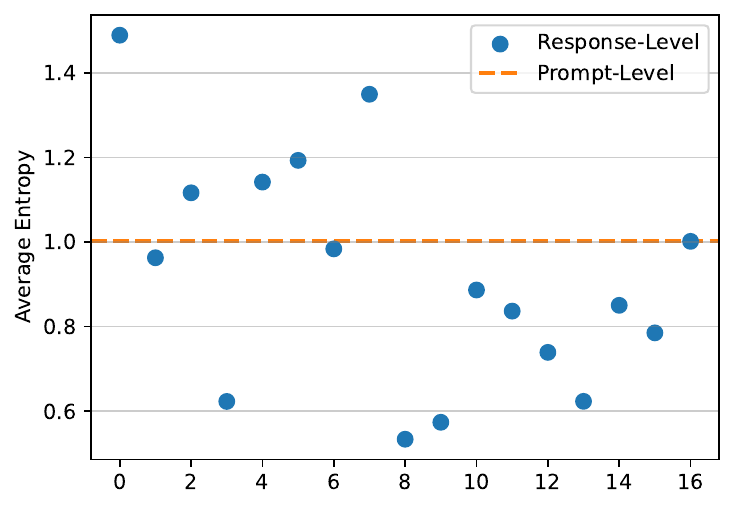}}\label{fig:response_avg_entropy}
\end{tabular}
\caption{Comparison of average entropy: (a) Prompt-level vs. batch-level across all prompts; (b) Response-level vs. prompt-level across all responses.}
\label{fig:entropy_variation}
\end{figure*}

\subsection{Intra-Response Entropy as a Zero-Shot Proxy for Token Roles}
\label{subsec:token_identification}

Prior RL approaches like GRPO~\citep{GRPO} and DAPO~\citep{DAPO} typically adopt a uniform token-level optimization strength to all output tokens. This undifferentiated treatment fails to account for the distinct functional roles that different tokens play in the reasoning process (e.g., factual recall vs. logical decision points). 
Recent work shows that RL-driven improvements in LLM reasoning stem mainly from enhancing logical behaviors such as reflection and planning, which \textbf{integrate existing model capabilities}, rather than directly modifying the model's factual memory or primitive skills~\citep{yue2025does, wen2025reinforcement}.
Thus, during RL training, tokens associated with \textit{factual knowledge} or \textit{base-level skills} should largely retain their original distributions, while tokens involved in \textit{logical reasoning and decision-making} require stronger learning signals and targeted exploration. To address this issue, a crucial first step is to identify critical reasoning tokens.

\paragraph{Entropy-based Token Identification.}
Recent work proposes entropy as an effective signal for identifying critical tokens, observing that high-entropy tokens frequently appear at logical transition points between reasoning segments~\citep{80/20}. In contrast, low-entropy tokens typically complete ongoing statements or syntactic structures. This observation aligns with our hypothesis that entropy discriminates between reasoning-oriented and knowledge-oriented tokens. To empirically verify this, we analyze token entropy distributions of 1024 responses (each prompt 16 times) generated by \DSOneB during training on mathematical tasks. Following~\citet{80/20}, we visualize the top-100 highest entropy tokens and the top-100 lowest entropy tokens and retain tokens that appear more than 100 times. The visualization in Figure~\ref{fig:wordcloud} shows that high-entropy tokens are mainly reasoning-related tokens, while most low-entropy tokens are related to factual knowledge or the suffix part of a word. These findings are also validated by recent studies~\citep{Lopti, cheng2025reasoning}. In summary, token entropy serves as an effective metric to distinguish between reasoning-oriented and knowledge-oriented tokens.

\subsubsection{Response-Level Entropy Statistics}
\label{subsubsec:response_level_entropy}

To distinguish token types, prior works compute token entropy quantiles or covariance statistics at the batch level~\citep{80/20, Entropy-Mechanism}. However, we find this suboptimal due to substantial entropy variation across responses from different prompts, as shown in Figure~\ref{fig:entropy_variation}. For instance, some prompts yield responses with average entropy far above/below the batch mean~(Figure~\ref{fig:entropy_variation}~(a)); even within a single prompt, entropy can vary across sampled responses significantly~(Figure~\ref{fig:entropy_variation}~(b)).

Therefore, batch-level statistics for token classification introduce a key drawback: if a response's overall entropy is low, even critical reasoning tokens may be misclassified as low-entropy, resulting in suboptimal training. For example, using the 80th percentile as a threshold can result in only 4.34\% of tokens being labeled as high-entropy in low-entropy responses. Conversely, for high-entropy responses, the proportion of high-entropy tokens may be abnormally inflated. To mitigate this, we adopt a \textbf{response-level} entropy statistics method for token classification, computing entropy quantiles independently within each response. Given a batch of $N$ rollout responses, let $e_t^i$ be the entropy of token $t$ in response $o^i$. We compute the $\rho$-quantile of token entropy for each response as a threshold:
\begin{equation}
\tau_\rho^i = \operatorname{Quantile} \Big( \{e_t^i\}_{t=1}^{|o^i|}, \rho \Big),
\label{eq:entropy_threshold}
\end{equation}
where $\rho \in (0, 1)$ denotes the quantile level (e.g., $\rho=0.8$ corresponds to the 80th percentile).

\subsubsection{Dual-Anchor Token-Conditioned Trust Region}
\label{subsubsec:our_method}

To address these issues, we propose a framework that performs synchronous updates while applying differentiated training constraints to different token types. Using response-level entropy as the criterion, we distinguish knowledge-type (low-entropy) from reasoning-type (high-entropy) tokens. Unlike prior works that adopt isolation strategies (e.g., gradient masking or asynchronous training), our method optimizes all tokens jointly but dynamically modulates the constraint intensity. We first introduce a token-level gating variable $\omega_t^i = \mathbb{I}(e_t^i \geq \tau_{\rho}^i) \in \{0,1\}$ based on the entropy quantile $\tau_{\rho}^i$ computed with Eq.~(\ref{eq:entropy_threshold}).

\paragraph{Local Proximal Anchor.}
To control the magnitude of policy updates at each step, we apply stricter clip ranges to knowledge-type (low-entropy) tokens to preserve the base model's capabilities and looser clip ranges to reasoning-type (high-entropy) tokens to encourage exploratory behavior. Given a batch of responses, we first compute the entropy quantile $\tau_\rho^i$ of token entropy within each response using~\eqref{eq:entropy_threshold}. Based on the computed entropy threshold, we categorize tokens into different types and assign distinct clipping ranges to each type accordingly:
\begin{equation}
\varepsilon(e_t^i) = \omega_t^i \varepsilon^\text{r} + (1-\omega_t^i) \varepsilon^\text{k}
\label{eq:clip}
\end{equation}

\paragraph{Global Reference-Policy Anchor.}
In RL training, the KL divergence penalty is commonly used to constrain the overall deviation of the trained policy from a reference policy~\citep{GRPO}. Although recent works~\citep{DAPO, Dr-GRPO, Open-Reasoner-Zero, GPG, VAPO, Skywork-OR1} advocate removing the KL divergence penalty, ProRL~\citep{ProRL} argues that this typically holds for base models without extensive SFT and using the KL penalty is crucial for training stability. Our experimental results also confirm that fully removing the KL penalty leads to training collapse and degraded performance, as shown in Section~\ref{subsubsec:ablation_kl}. Moreover, applying uniform KL penalties across all tokens, including high-entropy ones, significantly slows learning and reduces final performance.

Therefore, we extend the conventional KL penalty by adapting it based on the functional type of each token. Specifically, we apply a stronger KL penalty (i.e., a larger KL weight) to knowledge-type tokens (low entropy) to preserve the base model's factual knowledge. In contrast, we apply a weaker KL penalty (i.e., a smaller KL weight) to reasoning-type tokens (high entropy), enabling greater flexibility in critical reasoning regions. The coefficients of KL constraints are computed as:
\begin{equation}
\beta(e_t^i) = \omega_t^i \beta^\text{r} + (1-\omega_t^i) \beta^\text{k}
\label{eq:KL}
\end{equation}

Finally, the overall objective of our algorithm is formulated as follows:
\begin{equation}
\begin{aligned}
    \mathcal{J}_{\text{Archer}}(\theta) = &\ \mathbb{E}_{(q, a) \sim \mathcal{D}, \{o^i\}_{i=1}^G \sim \pi_{\theta_{\text{old}}}(\cdot \mid q)} \Bigg[ \frac{1}{\sum_{i=1}^G |o^i|} \sum_{i=1}^G \sum_{t=1}^{|o^i|} \\
    & \bigg( \min\left( r_t^i(\theta) \hat{A}_t^i, \operatorname{clip}\big( r_t^i(\theta), 1-\textcolor{red}{\varepsilon(e_t^i)}, 1+\textcolor{red}{\varepsilon(e_t^i)} \big) \hat{A}_t^i \right) - \textcolor{red}{\beta(e_t^i)} \mathbb{D}_{\text{KL}}(\pi_\theta \| \pi_{\text{ref}}) \bigg) \Bigg],
\end{aligned}
\label{eq:loss}
\end{equation}
where differentiated clipping and KL constraints are denoted using red color. The full algorithm of \ourmethod is shown in Algorithm~\ref{alg:method}.

\begin{algorithm}[!h]
\caption{\ourmethod}\label{alg:method}
\begin{algorithmic}[1]
\Require Base model $\pi_{\text{base}}$, prompt dataset $\mathcal{D}$, quantile level $\rho$, clipping thresholds $\varepsilon^\text{r}, \varepsilon^\text{k}$, KL coefficients $\beta^\text{r}, \beta^\text{k}$
\State Initialize policy model $\pi_\theta \leftarrow \pi_{\text{base}}$ and reference model $\pi_{\text{ref}} \leftarrow \pi_{\text{base}}$
\For{step $=1, 2, \ldots, T$}
\State Sample a batch of prompts $\mathcal{D}_b$ from $\mathcal{D}$
\State Generate responses $\{o^i\}_{i=1}^G$ for each prompt $q$ in the batch
\For{each response $o^i$}
    \State Compute the $\rho$-quantile of token entropy $\tau_\rho^i$ with~\eqref{eq:entropy_threshold}
    \State Compute clipping thresholds and coefficients of KL penalty with~\eqref{eq:clip} and~\eqref{eq:KL}
\EndFor
\State Update the policy model $\pi_\theta$ using~\eqref{eq:loss}
\EndFor
\end{algorithmic}
\end{algorithm}

\begin{figure*}[!t]
\centering
\includegraphics[width=1.0\textwidth]{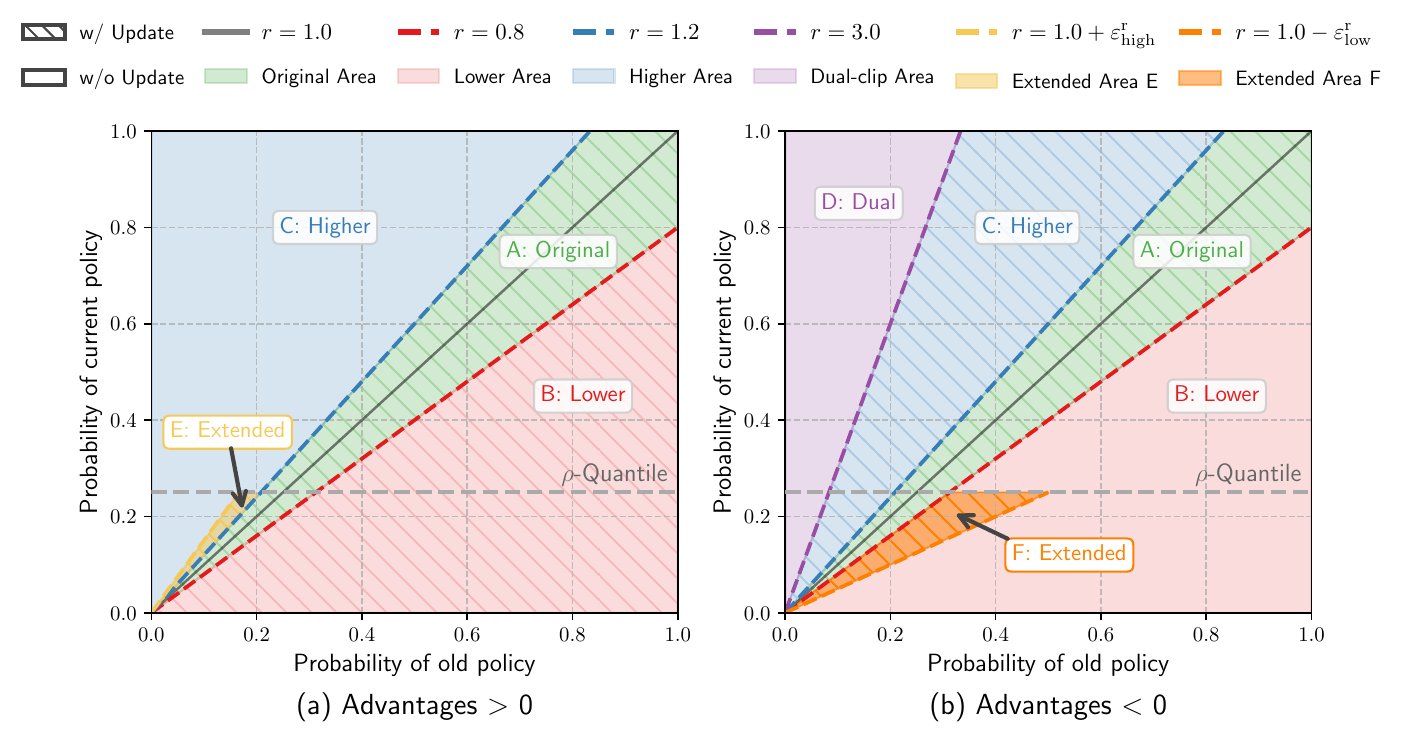}
\caption{Visualization of PPO clip regions. The x-axis shows the sampled probability of a specific token $\pi_{\theta_\text{old}}$ during generation, and the y-axis shows the probability of the token under the current policy $\pi_\theta$. Region A denotes the optimization area for original GRPO. Regions B and C are areas below and above the clipping threshold, respectively. Region D is the area for dual-clip~\citep{Dual-clip-PPO}. (a) When $\hat{A}_t^i > 0 $, \ourmethod optimizes region E. (b) When $\hat{A}_t^i < 0 $, \ourmethod optimizes region F.}
\label{fig:ppo_clip_regions}
\end{figure*}

\subsubsection{Visualization of RL Optimization Regions}

To better clarify the mechanism of our method, we visualize the optimization regions produced by the GRPO loss for different token types in Figure~\ref{fig:ppo_clip_regions}. Each data point in the coordinate system represents the importance sampling ratio $r_t^i$ between the current and old policy probabilities. Figure~\ref{fig:ppo_clip_regions}(a) shows tokens with positive advantage values ($\hat{A}_t^i > 0$), while Figure~\ref{fig:ppo_clip_regions}(b) shows tokens with negative advantages ($\hat{A}_t^i < 0$). The colored regions mark the areas divided by the clipping thresholds. The shaded areas (Regions A, B for $\hat{A}_t^i > 0$ and Regions A, C for $\hat{A}_t^i < 0$) indicate where GRPO updates the model. Our method \textbf{extends the clipping boundaries for high-entropy tokens}, which are typically low-probability but are important for reasoning. As shown in Figure~\ref{fig:ppo_clip_regions}, Regions E and F correspond to the \textbf{newly extended optimization areas} introduced by \ourmethod. Region E provides \textbf{additional reward signals} to high-entropy tokens when $\hat{A}_t^i > 0$, while Region F applies \textbf{stronger penalties} to high-entropy tokens when $\hat{A}_t^i < 0$. This design \textbf{increases the model's focus on learning reasoning-critical tokens}.

\section{Experiments}
\label{sec:experiments}

\subsection{Setup}
\label{subsec:exp_setup}

\paragraph{Models and Baselines.}

We adopt \DSOneB~\citep{DeepSeek-R1}, \QFourB, \QEightB, and \QThirtyB\footnote{We use non-thinking mode for Qwen3 models.}~\citep{Qwen3} as the base model and compare \ourmethod against the following methods:
(1) \textbf{Base Model}: The raw distilled model without further training.
(2) \textbf{GRPO}~\citep{GRPO}: A common RLVR algorithm combined with token-level loss~\citep{DAPO}.
For \DSOneB, we additionally compare:
(3) \textbf{DeepScaleR-1.5B}~\citep{DeepScaleR}: A 1.5B model trained on mathematical tasks with iterative context length expansion.
(4) \textbf{DeepCoder-1.5B}~\citep{DeepCoder}: A 1.5B model trained on code datasets, also utilizing context expansion strategies.
(5) \textbf{FastCuRL-1.5B-V3}~\citep{FastCuRL}: A strong 1.5B model with curriculum RL training.
(6) \textbf{Nemotron-1.5B}~\citep{ProRL}: Currently the best 1.5B reasoning model that RL-trained with \DSOneB as the base model.
For \QFourB and \QEightB, we also compare: (7) \textbf{80/20}~\citep{80/20} and (8) \textbf{AR-Lopti}~\citep{Lopti}.

\paragraph{Training Data.}

For code domain, we construct a high-quality code training dataset from three publicly available sources: DeepCoder~\citep{DeepCoder}, CodeContests~\citep{CodeContests}, and CodeForces~\citep{CodeForces}. Notably, CodeContests and CodeForces augment original problems with extensive test cases, which reduces false positives (i.e., incorrect solutions that pass test cases). Therefore, we prioritize these two datasets over DeepCoder in cases of duplication. After rigorous cleaning and filtering steps (detailed in Appendix~\ref{app:dataset}), we obtain a final corpus of 6,753 programming problems. For mathematics domain, we use datasets from DeepScaleR~\citep{DeepScaleR}, Skywork-OR1~\citep{Skywork-OR1}, and DAPO~\citep{DAPO}. We merge these datasets and apply N-gram overlap removal to eliminate duplicates. After additional verification and filtering steps (see Appendix~\ref{app:dataset}), we derive a final mathematics training set of 51,800 problems.

\paragraph{Evaluation and Metrics.}

We conduct evaluation on both mathematical and coding benchmarks. For mathematics, we use six challenging datasets: AIME24~\citep{AIME24}, AIME25~\citep{AIME25}, AMC23~\citep{AMC23}, MATH-500~\citep{PRM800K}, Minerva Math~\citep{Minerva-Math}, and OlympiadBench~\citep{OlympiadBench}. For coding, we adopt the widely used LiveCodeBench v5 (2024.08.01-2025.02.01) and v6 (2025.02.01-2025.05.01)~\citep{LiveCodeBench}, which emphasize reasoning-intensive code generation. We use \texttt{vLLM}~\citep{vLLM} with temperature set to 0.8, \texttt{top\_p} set to 1.0, and maximum output length set to 32,768 tokens for inference. Due to the high variance of the outputs from reasoning models, we report \textit{avg@K} (pass@1 performance averaged over K outputs) and \textit{pass@K} for each benchmark. For benchmarks with few samples (AIME24/25 and AMC23), we set a larger K=64. We use K=16 for LiveCodeBench v6, K=8 for LiveCodeBench v5 and Minerva, and K=4 for MATH-500 and OlympiadBench. To ensure accurate evaluation, we adopt the verification functions from both DeepScaleR and Math-Verify\footnote{https://github.com/huggingface/Math-Verify} for mathematics problems.

\paragraph{Implementation Details.}
We perform RL training using the \texttt{verl} framework~\citep{verl}. For GRPO-based baselines, we use clipping thresholds of $\varepsilon_{\text{low}} = 0.2$ and $\varepsilon_{\text{high}} = 0.28$. KL penalty loss and entropy regularization loss are omitted from the loss function. During training, we sample 16 rollouts per prompt with a temperature of 1.0. We set a maximum response length of 32,768 for \DSOneB, 8,192 for \QFourB and \QThirtyB, and 6,144 for \QEightB. The batch size is set to 64, the mini-batch size to 32, and the learning rate to $1 \times 10^{-6}$. For \ourmethod, we set $\rho=0.8$ following~\citet{80/20}. For clipping ranges and KL coefficients, we use $\varepsilon^\text{r}=0.5$, $\varepsilon^\text{k}=0.2$, $\beta^{\text{r}}=0.0$, and $\beta^{\text{k}}=0.001$. All experiments are conducted on 2 compute nodes, each equipped with 8 $\times$ NVIDIA H800 80GB GPUs.

\begin{table*}[!t]
\centering
\caption{Evaluation results on mathematical benchmarks. The results of \ourmethod are \colorbox{\mycolor}{shaded} and the highest values are \textbf{bolded}.}
\resizebox{1.0\textwidth}{!}{
\begin{tabular}{
L{3.2cm}   
C{0.65cm} C{0.75cm} @{\hspace{0.40cm}}  
C{0.65cm} C{0.75cm} @{\hspace{0.40cm}}  
C{0.65cm} C{0.75cm} @{\hspace{0.40cm}}  
C{0.65cm} C{0.75cm} @{\hspace{0.35cm}}  
C{0.65cm} C{0.75cm} @{\hspace{0.35cm}}  
C{0.65cm} C{0.75cm} @{\hspace{0.35cm}}  
C{0.7cm}   
}
\toprule
\multirow{2}{*}{\textbf{Method}} & \multicolumn{2}{c}{\textbf{AIME24}} & \multicolumn{2}{c}{\textbf{AIME25}} & \multicolumn{2}{c}{\textbf{AMC23}} & \multicolumn{2}{c}{\textbf{MATH-500}} & \multicolumn{2}{c}{\textbf{Minerva}} & \multicolumn{2}{c}{\textbf{Olympiad}} & \multirow{2}{*}{\textbf{Avg.}} \\
\cmidrule(lr){2-3} \cmidrule(lr){4-5} \cmidrule(lr){6-7} \cmidrule(lr){8-9} \cmidrule(lr){10-11} \cmidrule(lr){12-13}
\noalign{\vskip -0.85em}
& {\footnotesize\color{gray}{avg@64}} & {\footnotesize\color{gray}{pass@64}}
& {\footnotesize\color{gray}{avg@64}} & {\footnotesize\color{gray}{pass@64}}
& {\footnotesize\color{gray}{avg@64}} & {\footnotesize\color{gray}{pass@64}}
& {\footnotesize\color{gray}{avg@4}} & {\footnotesize\color{gray}{pass@4}}
& {\footnotesize\color{gray}{avg@8}} & {\footnotesize\color{gray}{pass@8}}
& {\footnotesize\color{gray}{avg@4}} & {\footnotesize\color{gray}{pass@4}}
& \\
\midrule
\textbf{DeepSeek-R1-1.5B} & 30.6 & 80.0 & 23.5 & 63.3 & 70.7 & \textbf{100.0} & 83.6 & 92.4 & 27.6 & 48.2 & 44.6 & 59.4 & 46.8 \\
$\drsh$ GRPO  & 42.1 & 80.0 & 28.6 & 56.7 & 80.3 & 97.5 & 87.6 & \textbf{94.6} & 29.2 & 46.3 & 53.2 & 65.8 & 53.5 \\
$\drsh$ DeepScaleR-1.5B   & 42.0 & \textbf{83.3} & 29.0 & 63.3 & 81.3 & \textbf{100.0} & 87.7 & 93.6 & 30.3 & \textbf{51.1} & 50.7 & 61.0 & 53.5 \\
$\drsh$ FastCuRL-1.5B-V3  & 48.1 & 80.0 & 32.7 & 60.0 & \textbf{86.4} & 95.0 & 89.8 & 94.0 & 33.6 & 50.0 & 55.3 & 64.3 & 57.7 \\
$\drsh$ Nemotron-1.5B     & 48.0 & 76.7 & 33.1 & 60.0 & 86.1 & 97.5 & 90.6 & 93.6 & 35.3 & 47.8 & 59.2 & 66.8 & 58.7 \\
\rowcolor{cyan!10} $\drsh$ Archer-Math-1.5B & \textbf{48.7} & \textbf{83.3} & \textbf{33.8} & \textbf{70.0} & 86.0 & 97.5 & \textbf{90.8} & 94.4 & \textbf{35.7} & \textbf{51.1} & \textbf{59.3} & \textbf{67.1} & \textbf{59.1} \\
\midrule
\textbf{\QFourB} & 23.6 & 56.7 & 18.3 & 63.3 & 67.7 & 95.0 & 84.5 & 92.4 & 41.5 & 56.3 & 54.1 & 66.6 & 48.3 \\ 
$\drsh$ GRPO     & 43.4 & \textbf{83.3} & 35.5 & \textbf{70.0} & 84.3 & \textbf{97.5} & 91.7 & 95.8 & 47.2 & 58.5 & 67.4 & 75.8 & 61.6 \\ 
$\drsh$ 80/20    & 50.4 & \textbf{83.3} & 40.5 & \textbf{70.0} & 88.9 & \textbf{97.5} & 93.6 & 97.2 & 47.2 & 57.0 & 69.3 & 78.0 & 65.0 \\
$\drsh$ AR-Lopti & 48.6 & 80.0 & 38.9 & \textbf{70.0} & 88.7 & \textbf{97.5} & 93.7 & \textbf{97.4} & 49.1 & 58.5 & 69.5 & 78.2 & 64.8 \\
\rowcolor{cyan!10} $\drsh$ \ourmethod-Math-4B & \textbf{51.4} & \textbf{83.3} & \textbf{43.1} & \textbf{70.0} & \textbf{91.0} & \textbf{97.5} & \textbf{95.1} & \textbf{97.4} & \textbf{51.2} & \textbf{60.7} & \textbf{71.6} & \textbf{79.4} & \textbf{67.1} \\
\midrule
\textbf{\QEightB} & 27.0 & 63.3 & 19.1 & 56.7 & 68.9 & \textbf{97.5} & 83.6 & 92.4 & 43.9 & 58.1 & 55.7 & 69.6 & 49.7 \\ 
$\drsh$ GRPO      & 50.3 & \textbf{83.3} & 34.1 & 66.7 & 84.1 & 95.0 & 92.7 & 96.0 & 50.2 & 61.4 & 68.2 & 76.1 & 63.3 \\ 
$\drsh$ 80/20    & 50.9 & 80.0 & 40.1 & 70.0 & 87.3 & \textbf{97.5} & 94.1 & 96.8 & 49.2 & 59.6 & \textbf{70.0} & \textbf{78.5} & 65.3 \\
$\drsh$ AR-Lopti & 50.6 & \textbf{83.3} & 38.1 & 60.0 & 88.4 & \textbf{97.5} & 93.3 & 97.4 & 49.6 & 58.8 & 68.9 & 78.3 & 64.8 \\
\rowcolor{cyan!10} $\drsh$ \ourmethod-Math-8B & \textbf{56.1} & 76.7 & \textbf{43.6} & \textbf{73.3} & \textbf{91.8} & \textbf{97.5} & \textbf{94.8} & \textbf{97.8} & \textbf{51.5} & \textbf{62.1} & 69.4 & 77.3 & \textbf{67.9} \\ 
\midrule
\textbf{\QThirtyB} & 30.1 & 70.0 & 19.9 & 56.7 & 74.3 & \textbf{100.0} & 88.4 & 96.0 & 47.7 & 59.6 & 59.6 & 71.2 & 53.3 \\ 
$\drsh$ GRPO       & 59.5 & 80.0 & 43.1 & 73.3 & 91.9 & 97.5 & 95.3 & 97.8 & 51.9 & 61.8 & 71.0 & 79.2 & 68.8 \\ 
\rowcolor{cyan!10} $\drsh$ \ourmethod-Math-30B & \textbf{70.3} & \textbf{90.0} & \textbf{53.9} & \textbf{86.7} & \textbf{94.4} & 97.5 & \textbf{96.4} & \textbf{99.0} & \textbf{55.5} & \textbf{64.7} & \textbf{77.4} & \textbf{84.0} & \textbf{74.6} \\ 
\bottomrule
\end{tabular}%
}
\label{tab:main_math}%
\end{table*}%

\begin{table*}[!h]
\centering
\caption{Evaluation results on code benchmarks. The results of \ourmethod are \colorbox{\mycolor}{shaded} and the highest values are \textbf{bolded}.}
\resizebox{0.90\textwidth}{!}{
\begin{tabular}{lccccccc}
\toprule
\multirow{2}[2]{*}{\textbf{Method}} & \multicolumn{2}{c}{\textbf{LCB v5 (2024.08.01-2025.02.01)}} & \multicolumn{2}{c}{\textbf{LCB v6 (2025.02.01-2025.05.01)}} & \multirow{2}[2]{*}{\textbf{Avg.}} \\
\cmidrule(lr){2-3} \cmidrule(lr){4-5}
  & \multicolumn{1}{c}{\color{gray}{avg@8}}
  & \multicolumn{1}{c}{\color{gray}{pass@8}}
  & \multicolumn{1}{c}{\color{gray}{avg@16}}
  & \multicolumn{1}{c}{\color{gray}{pass@16}}
  & \\
\midrule
\textbf{DeepSeek-R1-1.5B} & 16.7 & 29.0 & 17.2 & 34.4 & 17.0 \\
$\drsh$ GRPO  & 26.0 & 40.5 & 27.6 & 43.5 & 26.8 \\
$\drsh$ DeepCoder-1.5B  & 23.3 & 39.1 & 22.6 & 42.0 & 23.0 \\
$\drsh$ Nemotron-1.5B   & 26.1 & 35.5 & 29.5 & 42.8 & 27.8 \\
\rowcolor{cyan!10} $\drsh$ Archer-Code-1.5B & \textbf{29.4} & \textbf{43.7} & \textbf{30.2} & \textbf{45.8} & \textbf{29.8} \\
\midrule
\textbf{\QFourB} & 23.8 & 35.8 & 24.0 & 35.1 & 23.9 \\
$\drsh$ GRPO & 40.8 & 55.2 & 36.6 & 45.0 & 38.7 \\
\rowcolor{cyan!10} $\drsh$ \ourmethod-Code-4B & \textbf{42.1} & \textbf{57.0} & \textbf{37.5} & \textbf{48.1} & \textbf{39.8} \\
\bottomrule
\end{tabular}%
}
\label{tab:main_code}%
\end{table*}%

\subsection{Main Results}
\label{subsec:exp_main}

\paragraph{Comparison with Base Model and GRPO.}
The results in Table~\ref{tab:main_math} and~\ref{tab:main_code} show that our dual-token constraint training strategy leads to significant improvements on both mathematical and coding tasks. Compared to the original base model (\DSOneB), the average accuracy increases by \textbf{18.1\%} on AIME24 and \textbf{10.3\%} on AIME25, resulting in an average gain of \textbf{12.3\%}. On coding benchmarks, the accuracy rises by \textbf{12.7\%} on LiveCodeBench v5 and \textbf{13.0\%} on LiveCodeBench v6. When applying our method upon GRPO, the performance consistently exceeds that of GRPO across all benchmarks, with average gains of \textbf{5.6\%} and \textbf{3.0\%} for mathematical and coding tasks, respectively. When applying \ourmethod to \QFourB and \QEightB, \ourmethod still outperforms GRPO by a large margin.
These results demonstrate the effectiveness of our optimization approach.

\paragraph{Comparison with SOTA Reasoning Models.}
We also compare \ourmethod with SOTA reasoning models trained with RL using \DSOneB as the base model. For coding tasks, our approach outperforms all comparable models, including the programming-specialized \textbf{DeepCoder-1.5B} and the general-purpose \textbf{Nemotron-1.5B}. On mathematical reasoning, our model achieves the highest average accuracy, surpassing both math-specialized models (\textbf{DeepScaleR-1.5B}) and \textbf{Nemotron-1.5B}. We report the training costs of \ourmethod and these open-source reasoning models, including the number of training steps, stages, and GPU hours in Table~\ref{tab:efficiency}. Notably, our model achieves the best results with only \textbf{single-stage} training and \textbf{fewer} GPU hours, without the complex multi-round training used by the other methods.
In addition to improvements in \textbf{pass@1}, our model also shows advantages in \textbf{pass@K} metrics, which suggests stronger reasoning diversity and higher capability limits of our method.

\subsection{Analysis}
\label{subsec:analysis}

\subsubsection{Impact of Masking Low-Entropy Tokens}
\label{subsubsec:ablation_masking}

\begin{wraptable}[9]{r}{0.40\textwidth}
\centering
\vspace{-0.9em}
\caption{\small Controlled masking ablation on \QFourB. Avg. is the mean of avg@K over the six mathematical benchmarks.}
\begin{tabular}{lc}
\toprule
\textbf{Method} & \textbf{Avg.} \\
\midrule
\ourmethod (masking) & 66.0 \\
\ourmethod & \textbf{67.1} \\
\bottomrule
\end{tabular}
\label{tab:masking_ablation}
\end{wraptable}
To isolate the effect of joint optimization, we compare \ourmethod with a masking variant on \QFourB across the six mathematical benchmarks. Both variants use the same response-level entropy criterion, training setup, and hyperparameters. The only difference is that the masking variant masks the gradients of low-entropy tokens. As shown in Table~\ref{tab:masking_ablation}, \ourmethod improves the average accuracy from 66.0 to 67.1, a gain of 1.1 percentage points. This result suggests that retaining the training signals of low-entropy tokens with differentiated constraints is more effective than masking them in this setting, supporting joint optimization across token types.

\subsubsection{Impact of Different Entropy Thresholds}

We further ablate the entropy quantile threshold $\rho$ used to identify high-entropy tokens in Eq.~\eqref{eq:entropy_threshold}. A larger $\rho$ selects fewer tokens as reasoning-type tokens, making the criterion more conservative, while a smaller $\rho$ allows more tokens to receive relaxed clipping and weaker KL constraints.

\begin{table*}[!h]
\centering
\caption{Ablation of the entropy quantile threshold $\rho$ on mathematical benchmarks. The highest values are \textbf{bolded}.}
\resizebox{0.85\textwidth}{!}{
\begin{tabular}{lccccccc}
\toprule
{$\bm{\rho}$} 
& \textbf{AIME24} 
& \textbf{AIME25} 
& \textbf{AMC23} 
& \textbf{MATH-500} 
& \textbf{Minerva} 
& \textbf{Olympiad} 
& {\textbf{Avg.}} \\
\midrule
0.7  & 50.7 & 42.6 & \textbf{91.2} & \textbf{95.3} & 50.6 & 71.0 & 66.9 \\
0.8  & \textbf{51.4} & \textbf{43.1} & 91.0 & 95.1 & \textbf{51.2} & \textbf{71.6} & \textbf{67.1} \\
0.9  & 48.3 & 39.9 & 90.9 & 95.0 & 50.4 & 70.3 & 65.8 \\
\bottomrule
\end{tabular}%
}
\label{tab:abla_thresh}%
\end{table*}%
\FloatBarrier

As shown in Table~\ref{tab:abla_thresh}, $\rho=0.8$ achieves the best average performance, while the stricter threshold $\rho=0.9$ consistently degrades results. This suggests that an overly conservative threshold misclassifies many potentially reasoning-critical tokens as low-entropy tokens, thereby restricting their updates with tighter clipping and stronger KL regularization. In contrast, using a moderately smaller threshold exposes more high-entropy tokens to exploratory updates, which provides richer learning signals and leads to better model performance.

\subsubsection{Impact of Different KL Weights} 
\label{subsubsec:ablation_kl}

\begin{wraptable}[9]{r}{0.45\textwidth}
\centering
\vspace{-0.9em}
\caption{\small LiveCodeBench v5 performance with varying KL weights on low-entropy tokens.}
\resizebox{\linewidth}{!}{
\begin{tabular}{lc}
\toprule
\textbf{KL Weight} & \textbf{LiveCodeBench v5 (avg@8)} \\
\midrule
0.0 & 26.6 \\
0.001 & 29.4 \\
0.005 & 26.2 \\
\bottomrule
\end{tabular}
}
\label{tab:kl_weights_impact}
\end{wraptable}

We next vary the KL weight on low-entropy tokens and use the average n-gram repetition ratio as a proxy for collapse. Table~\ref{tab:kl_weights_impact} and Figure~\ref{fig:ablation_kl_weight} show that both removing KL regularization and using an overly large weight hurt performance. Without KL, entropy drops quickly and repetition rises, leading to unstable training and lower final accuracy. With a large KL weight, the model better preserves the base policy but learns more slowly.

In summary, both too little and too much KL regularization hurt the final model quality. Insufficient weighting accelerates learning but makes collapse more likely, which ends up reducing performance. In contrast, excessive weighting limits learning on low-entropy tokens and thus restricts the model's capabilities. These results \textbf{highlight the need for KL regularization on low-entropy tokens} to keep the model close to the base policy, which helps prevent collapse and retain key abilities. These observations further support our view that low-entropy tokens should be included in training, as masking them negatively affects overall learning.

\subsubsection{Impact of Clip Ranges on Different Token Types}
\label{subsubsec:ablation_clip}

\begin{wraptable}[15]{R}{0.45\textwidth}
\centering
\vspace{-0.2em}
\caption{\small LiveCodeBench v5 performance under different low-/high-entropy token clip thresholds.}
\resizebox{\linewidth}{!}{
\begin{tabular}{ccc}
\toprule
$\varepsilon^\text{k}$ & $\varepsilon^\text{r}$ & \textbf{LiveCodeBench v5 (avg@8)} \\
\midrule
\multicolumn{3}{c}{\textit{Varying Low-Entropy Token Clip}} \\
\midrule
0.1 & 0.4 & 24.6 \\
0.2 & 0.4 & \textbf{28.7} \\
0.3 & 0.4 & 26.0 \\
\midrule
\multicolumn{3}{c}{\textit{Varying High-Entropy Token Clip}} \\
\midrule
0.2 & 0.2 & 27.7 \\
0.2 & 0.4 & 28.7 \\
0.2 & 0.5 & \textbf{29.4} \\
0.2 & 0.6 & 26.0 \\
\bottomrule
\end{tabular}
}
\label{tab:clip_thresholds}
\end{wraptable}
We introduce different clip thresholds for different token types in~\eqref{eq:clip}. To investigate how the thresholds influence model performance, we vary the clip ranges for both high-entropy ($\varepsilon^\text{r}$) and low-entropy tokens ($\varepsilon^\text{k}$) and the results are shown in Table~\ref{tab:clip_thresholds}, Figure~\ref{fig:ablation_low_entropy_clip}, and~\ref{fig:ablation_high_entropy_clip}.


\paragraph{Different Low-Entropy Token Clip Thresholds.}
As shown in Figure~\ref{fig:ablation_low_entropy_clip}, we observe that increasing the clip threshold for low-entropy tokens produces effects similar to reducing their KL penalty weight: the model's entropy decreases more rapidly, which leads to faster learning and earlier performance improvements. However, this also causes the repetition ratio to rise more quickly, making the model more susceptible to overfitting or collapse, which harms final performance.

On the other hand, lowering the clip threshold for low-entropy tokens has effects similar to increasing their KL weight: improvements on LiveCodeBench v5 are slower and tend to converge a lower level. Interestingly, we observe an counterintuitive entropy dynamic during training. Instead of a consistently slow decline, as seen with higher KL weights, entropy initially drops sharply, then plateaus and remains stable.

These results indicate that adjusting the clip threshold for low-entropy tokens strongly affects both the training process and the final model performance. In contrast, the model is much less sensitive to changes in the clip threshold for high-entropy tokens.

\paragraph{Different High-Entropy Token Clip Thresholds.}
As illustrated in Figure~\ref{fig:ablation_high_entropy_clip}, increasing the clip threshold for high-entropy tokens encourages more exploration in the model's reasoning. This leads to a slightly faster reduction in entropy during training and can improve the performance. However, these differences become more noticeable mainly in the later stages of training. In the early stages, training dynamics and LiveCodeBench v5 performance show little difference across various high-entropy clip values.

\section{Conclusion}
\label{sec:conclusion}

In this work, we propose an entropy-aware, synchronized training framework that updates all tokens simultaneously while applying different regularization and clipping strategies depending on the type of token. By encouraging exploration on reasoning-related tokens and preserving factual correctness for knowledge-related tokens, our method balances the goals of keeping factual accuracy and improving logical reasoning. Extensive experiments on mathematical and code reasoning benchmarks show that our approach improves over the base model and outperforms existing SOTA models. These results indicate that coordinating the learning processes of different token types through entropy-aware constraints improves the reasoning abilities of LLMs. We believe this work highlights the interaction between factual knowledge and reasoning processes during RL training of LLMs, and suggests future research directions for fine-grained optimization methods that coordinate training signals across token types.

\paragraph{Limitations.}
Although \ourmethod significantly outperforms GRPO, it still has several limitations. First, entropy is only a heuristic proxy for token roles. Second, our current method uses a binary token partition with fixed clipping and KL coefficients. Future work could explore continuous constraints or more fine-grained token types.

\bibliography{main}
\bibliographystyle{plainnat}

\newpage
\appendix

\section{Experimental Details}
\label{app:experimental_details}

\subsection{Dataset}
\label{app:dataset}

\subsubsection{Code Domain}

\paragraph{Data Sources and Integration.}
The code dataset is compiled from three publicly available sources: DeepCoder, CodeContests, and CodeForces. Notably, CodeContests and CodeForces extend their original problem sets with a larger number of test cases, improving the reliability of evaluation and reducing the incidence of false positives—i.e., incorrect code that inadvertently passes tests. As such, these two datasets are prioritized. In cases of duplication with DeepCoder, we retain the entries from either CodeContests or CodeForces.

\paragraph{Data Cleaning and Filtering Pipeline.}
We apply a rigorous multi-stage cleaning and selection process to ensure dataset quality:

\begin{enumerate}
    \item \textbf{Test Case Preprocessing:} We remove illustrative test cases embedded in problem descriptions and discard problems with fewer than five test cases, which are more susceptible to false positives.
    
    \item \textbf{Model Validation and Difficulty Filtering:} Each problem is evaluated using 8-sample generation with a strong language model (Qwen3-30B-A3B~\citep{Qwen3}). We exclude problems for which all samples fail verification, filtering out flawed questions (e.g., with invalid test cases), overly long I/O problems beyond the verifier's capacity, or those that are excessively difficult—even for strong models. This reduces potential false negatives.
    
    \item \textbf{Problem Deduplication:} We perform N-gram-level deduplication to eliminate duplicate questions within the training corpus.
    
    \item \textbf{Test Set Contamination Prevention:} To prevent data leakage, we remove any overlapping problems by conducting N-gram-level deduplication against the evaluation set of LiveCodeBench v5.
    
    \item \textbf{Sampling Stability Filtering:} Using a warm-start model (\DSOneB), we generate 8 additional samples per problem. We remove problems where all generations are either completely correct or completely incorrect, thereby ensuring sufficient learning signal and gradient diversity.
\end{enumerate}

\paragraph{Data Standardization.}
All retained code problems are reformatted into either \textit{function-call} or \textit{stdin/stdout} formats, enabling consistent and automated validation via a code verifier.

\paragraph{Final Dataset.}
Following the aforementioned pipeline, we construct a high-quality code training dataset consisting of \textbf{6,753} problems.

\subsubsection{Mathematics Domain}

\paragraph{Data Sources and Integration.}
For the mathematics domain, we leverage existing curated datasets rather than raw symbolic corpora such as NuminaMath~\citep{NuminaMath}. Specifically, we integrate three high-quality, verifiable datasets: DeepScaleR, Skywork-OR1, and DAPO. The datasets are merged and deduplicated using N-gram overlap removal to eliminate redundancy.

\paragraph{Data Cleaning and Filtering Pipeline.}

\begin{enumerate}
    \item \textbf{Model Validation and Filtering:} Each math problem undergoes 8-sample generation using the Qwen3-30B-A3B model, followed by verification using a mathematical logic verifier. Problems for which all samples fail are excluded to remove noise, overly complex items, or verification bottlenecks that might cause false negatives.
    \item \textbf{Sampling Stability Filtering:} We repeat the 8-sample generation process using a warm-start model (\DSOneB) and discard problems with homogeneous sampling outcomes (i.e., all correct or all incorrect).
    \item \textbf{Test Set Contamination Prevention:} To avoid contamination of evaluation benchmarks, we perform N-gram deduplication against the AMC competition datasets (AIME24 and AIME25), ensuring zero overlap.
\end{enumerate}

\paragraph{Final Dataset.}
After rigorous verification and filtering, we obtain a final mathematics training corpus comprising approximately \textbf{51,800} high-quality problems suitable for reinforcement learning.

\section{Additional Experimental Results}
\label{app:exp_results}

\subsection{Efficiency}

\begin{table*}[!h]
\centering
\caption{Computational efficiency comparison between \ourmethod and the baselines on 1.5B models.}
\resizebox{0.60\textwidth}{!}{
\begin{tabular}{lccc}
\toprule
\textbf{Method} & \textbf{Training Steps} & \textbf{Stages} & \textbf{GPU Hours} \\
\midrule
\multicolumn{4}{c}{\textit{Math RL}} \\
\midrule
DeepScaleR-1.5B   & 1750 & 3 & 3,800 A100  \\
FastCuRL-1.5B-V3  & 2620 & 5 & — \\
Nemotron-1.5B     & 2500 & 8 & 16,000 H100 \\
\rowcolor{cyan!10} Archer-Math-1.5B  & 520  & 1 & 1,900 H800  \\
\midrule
\multicolumn{4}{c}{\textit{Code RL}} \\
\midrule
DeepCoder-1.5B   & — & — & — \\
Nemotron-1.5B    & 2500 & 8 & 16,000 H100 \\
\rowcolor{cyan!10} Archer-Code-1.5B & 320  & 1 & 1,000 H800  \\
\bottomrule
\end{tabular}%
}
\label{tab:efficiency}%
\end{table*}%

\subsection{Training Dynamics}

\begin{figure*}[!h]
\centering
\includegraphics[width=1\linewidth]{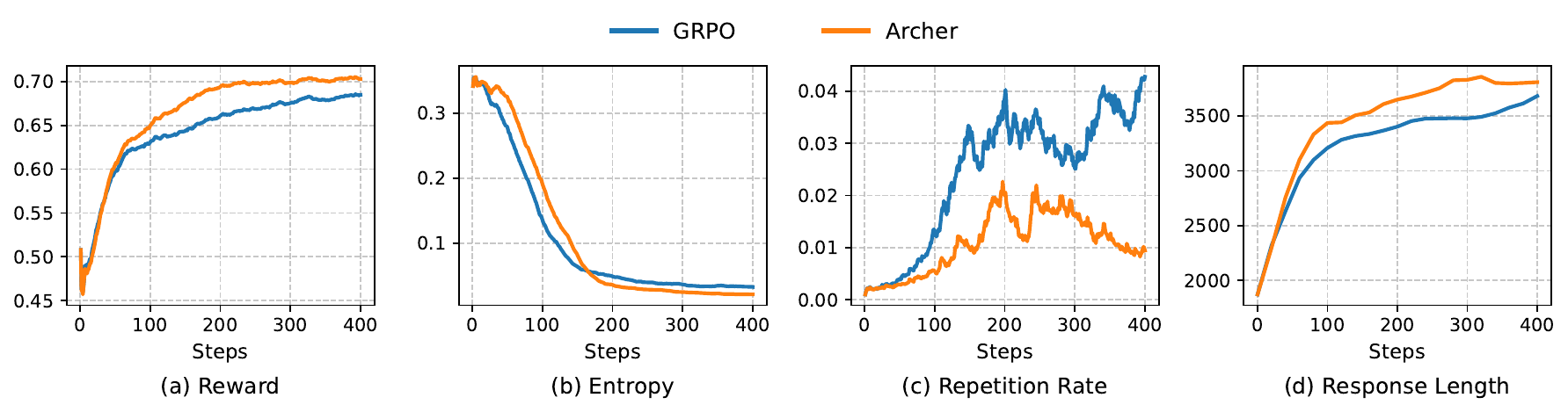}
\caption{The training dynamics curves of all methods on \QFourB, including (a) training reward, (b) model entropy, (c) repetition rate, and (d) response length. The curves are smoothed with EMA for better visualization.}
\label{fig:training_dynamics_qwen3_4b}
\end{figure*}

\begin{figure*}[!h]
\centering
\includegraphics[width=1\linewidth]{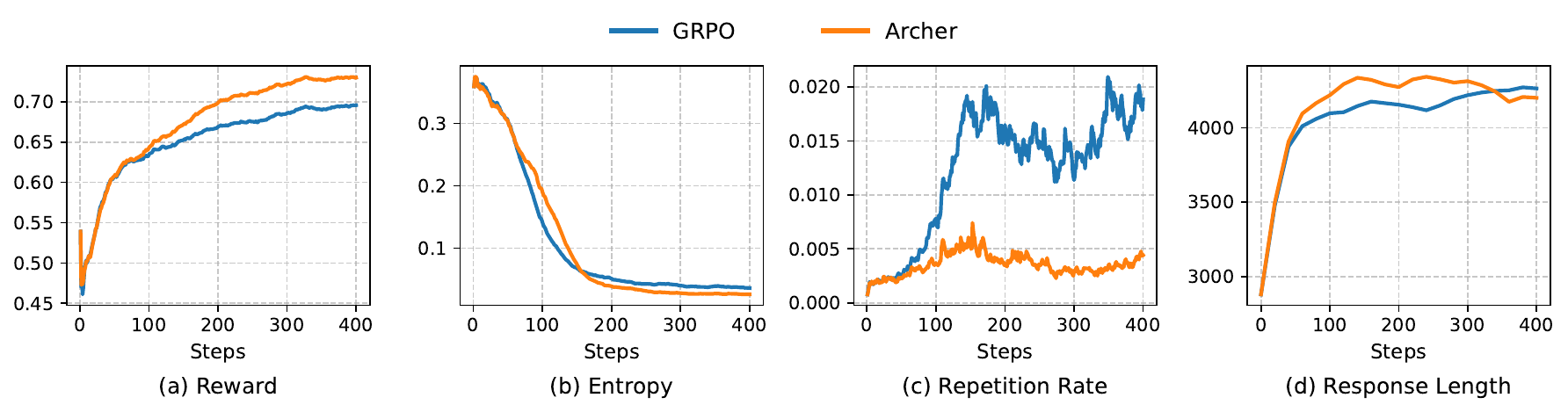}
\caption{The training dynamics curves of all methods on \QEightB, including (a) training reward, (b) model entropy, (c) repetition rate, and (d) response length. The curves are smoothed with EMA for better visualization.}
\label{fig:training_dynamics_qwen3_8b}
\end{figure*}

\begin{figure*}[!h]
\centering
\includegraphics[width=1\linewidth]{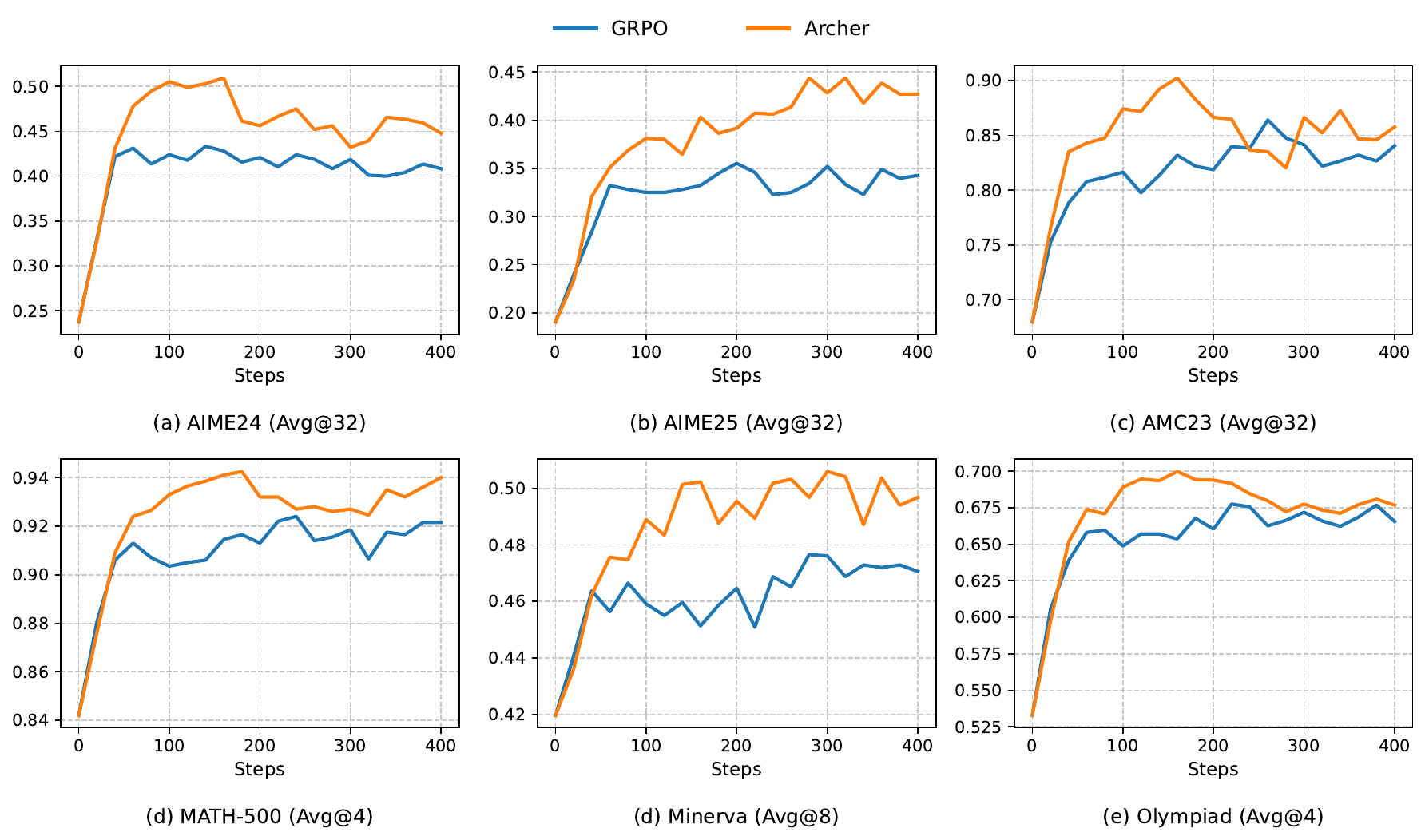}
\caption{The test curves of all methods trained with \QFourB on six mathematical benchmarks.}
\label{fig:all_test_curves_qwen3_4b}
\end{figure*}

We provide training dynamics curves and test curves on mathematical tasks in Figure~\ref{fig:training_dynamics_qwen3_4b},~\ref{fig:training_dynamics_qwen3_8b}, and~\ref{fig:all_test_curves_qwen3_4b}.
Figure~\ref{fig:training_dynamics_qwen3_4b} and~\ref{fig:training_dynamics_qwen3_8b} show that \ourmethod outperforms GRPO with lower repetition rate.

\begin{figure*}[!h]
\centering
\includegraphics[width=1.0\linewidth]{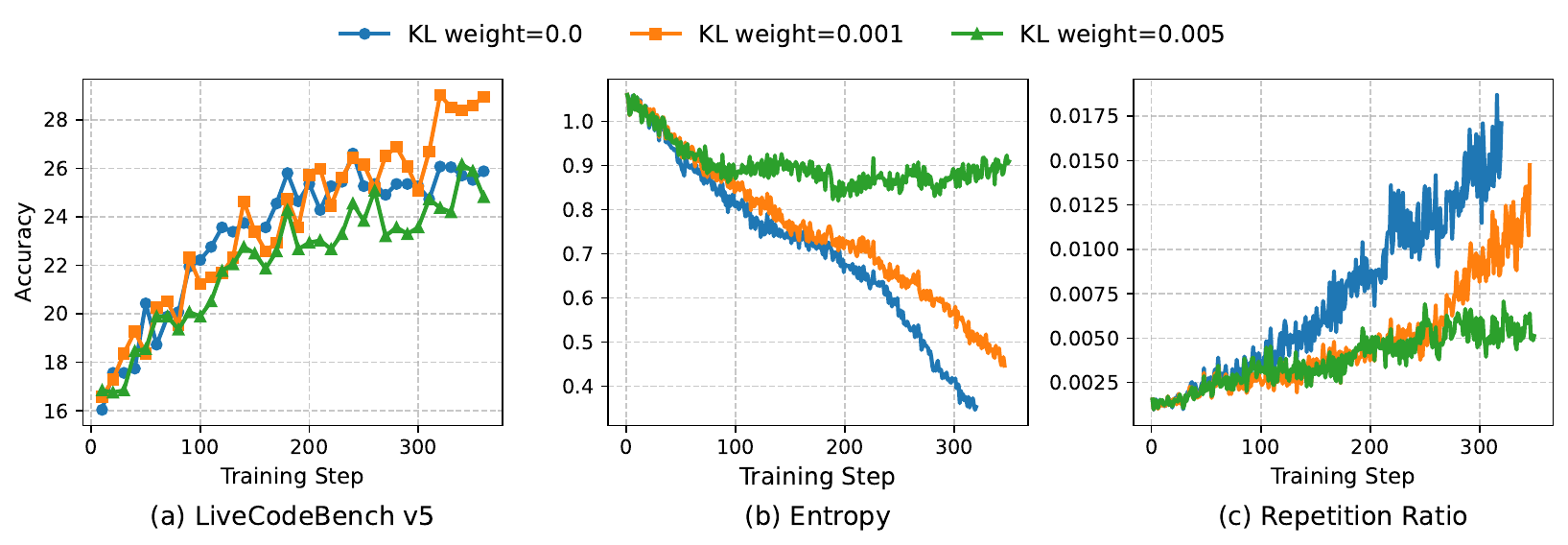}
\caption{Effects of varying KL weights on (a) model performance on LiveCodeBench v5, (b) model entropy, and (c) repetition ratio.}
\label{fig:ablation_kl_weight}
\end{figure*}

\begin{figure*}[!h]
\centering
\includegraphics[width=1.0\linewidth]{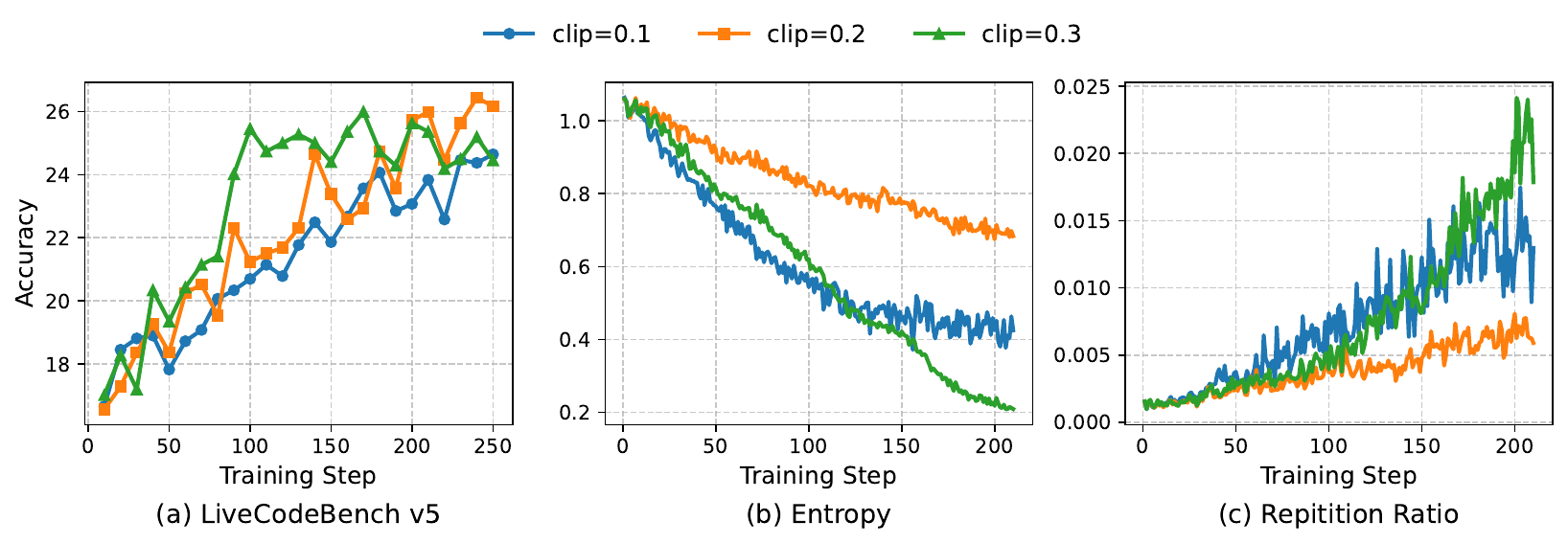}
\caption{Effects of varying the clip threshold of low-entropy tokens on (a) model performance on LiveCodeBench v5, (b) model entropy, and (c) repetition ratio.}
\label{fig:ablation_low_entropy_clip}
\end{figure*}

\subsection{Impact of Clip Ranges on High-Entropy Tokens}

\begin{figure*}[!h]
\centering
\includegraphics[width=1.0\linewidth]{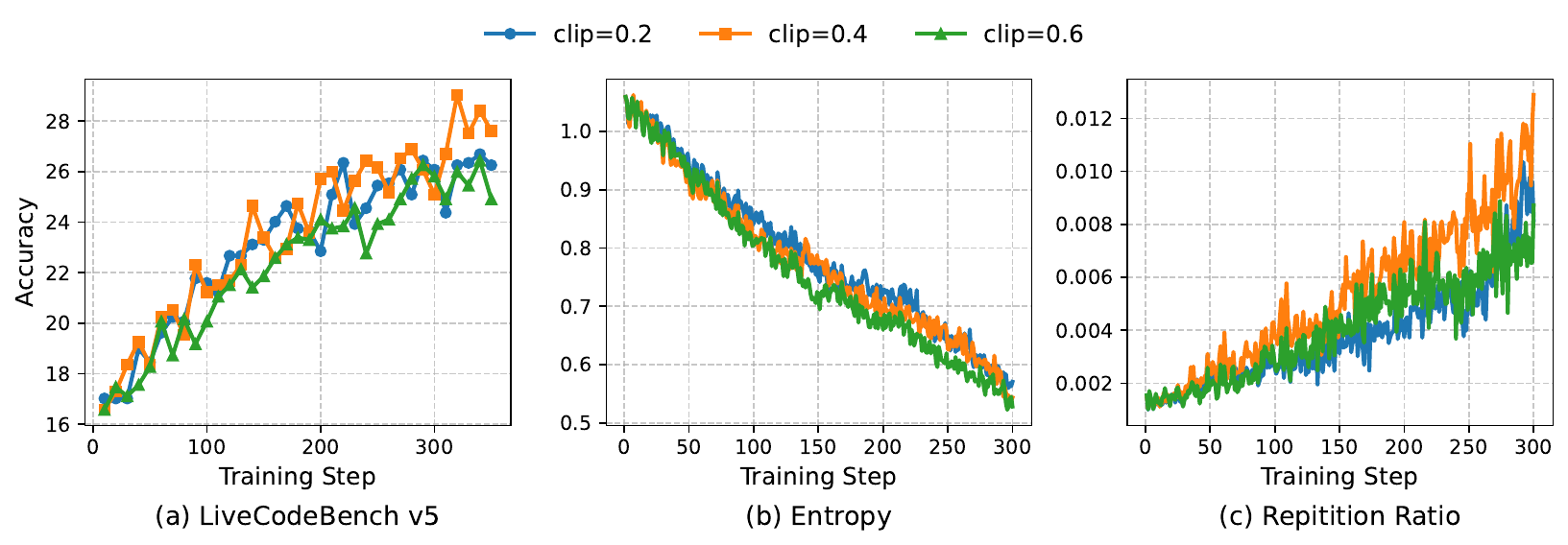}
\caption{Effects of varying clip value on high-entropy tokens on (a) model performance on LiveCodeBench v5, (b) model entropy, and (c) repetition ratio.}
\label{fig:ablation_high_entropy_clip}
\end{figure*}

\subsection{Mutual Enhancement Between Math RL and Code RL}

\begin{figure*}[!h]
\centering
\includegraphics[width=1.0\textwidth]{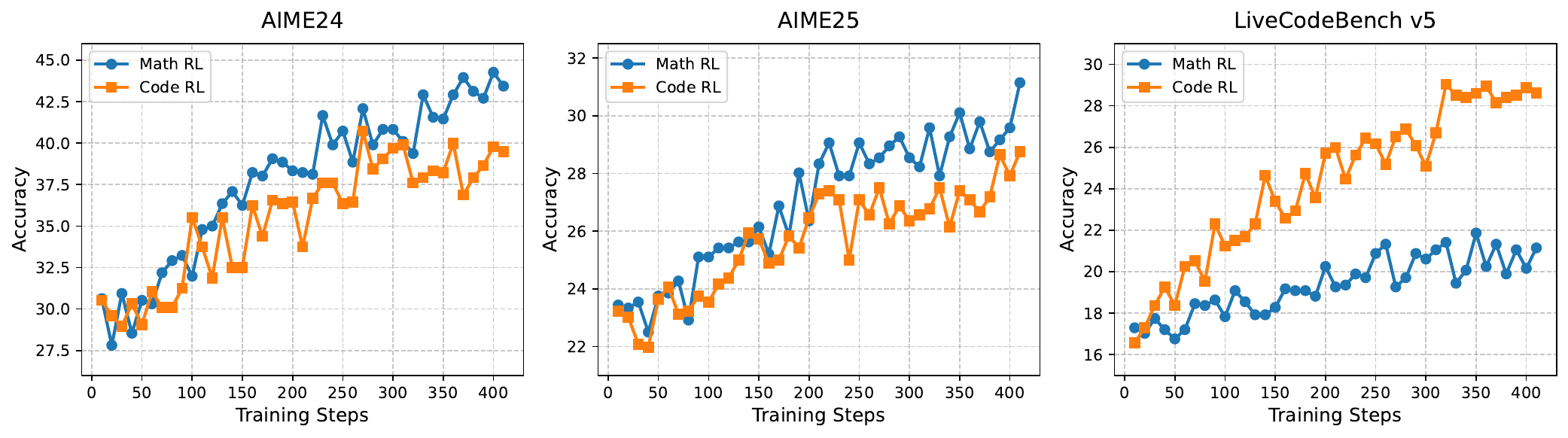}
\caption{Model performance on AIME24, AIME25, and LiveCodeBench v5 of math RL and code RL.}
\label{fig:math_code_rl}
\end{figure*}

Figure~\ref{fig:math_code_rl} shows results on AIME24, AIME25, and LiveCodeBench v5, comparing RL applied to math tasks (math RL) and code tasks (code RL). We observe that RL training in either domain leads to significant performance improvements not only in-domain but also on out-of-domain (OOD) benchmarks.

\begin{figure*}[!h]
\centering
\includegraphics[width=1.0\textwidth]{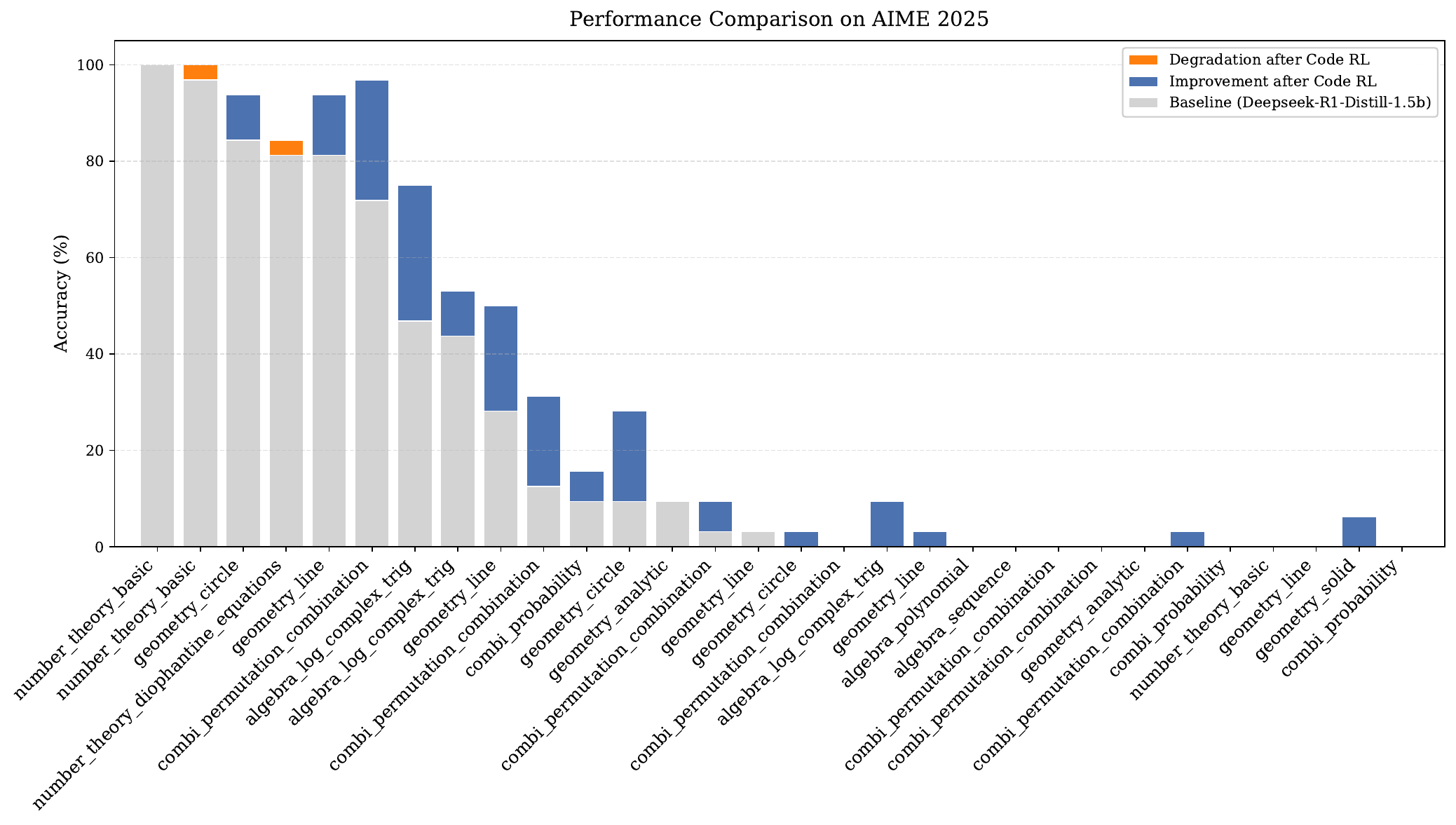}
\caption{Problem-level accuracy comparison between the base model and RL-trained model.}
\label{fig:problem_accuracy_comparison}
\end{figure*}

\begin{figure*}[!h]
\centering
\includegraphics[width=1.0\textwidth]{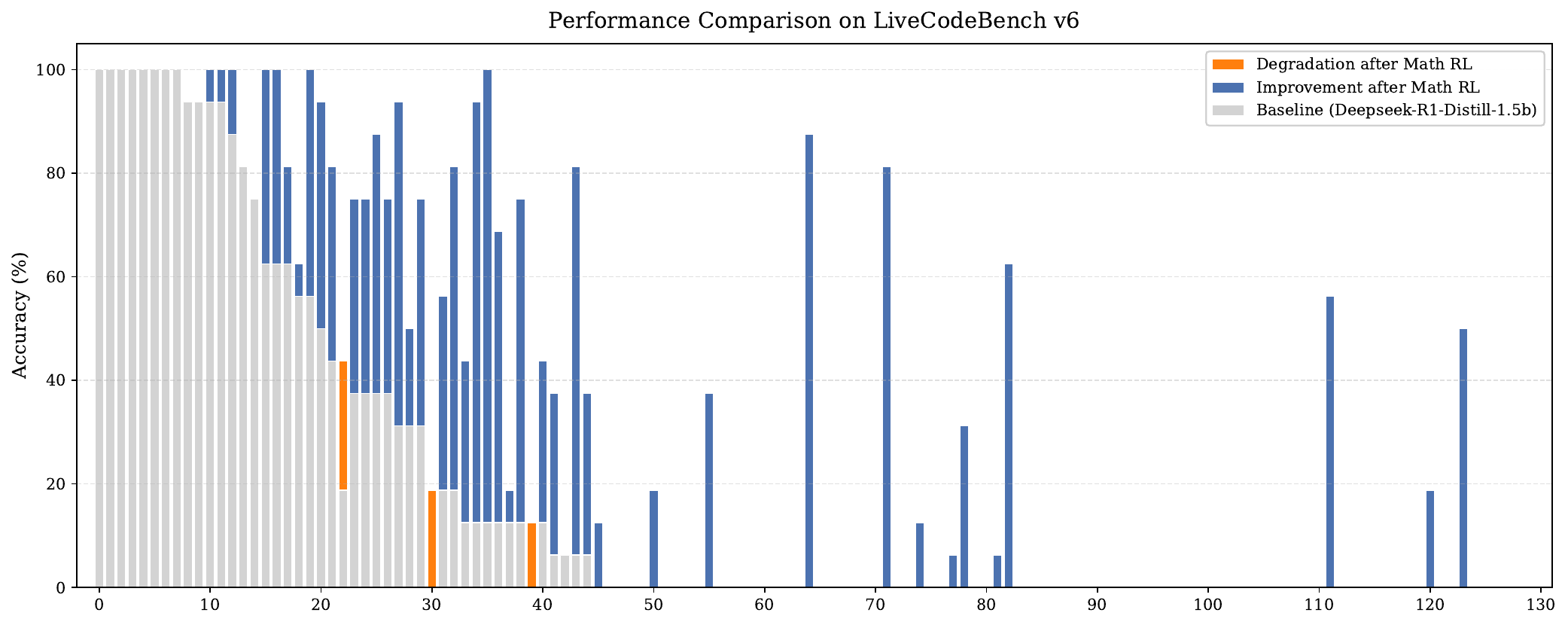}
\caption{Problem-level accuracy comparison on LiveCodeBench v6 between the base model and Math RL trained model.}
\label{fig:lcbv6_performance_comparison}
\end{figure*}

To analyze the source of these cross-domain improvements, we evaluate the base model and its math/code RL variants on OOD benchmarks (LiveCodeBench v5 and AIME24/25), measuring problem-level accuracy across all tasks. Unlike AceReason-Nemotron~\citep{AceReason-Nemotron}, which attributes the benefits of math RL on code tasks primarily to the presence of math-related subdomains (e.g., Algebra, Counting, Combinatorics), our results suggest a different explanation: performance improvements correlate more strongly with the intrinsic difficulty of the problems rather than their topical categories. Specifically, problems where the base model already achieves relatively high accuracy tend to benefit most from RL training, as shown in Figure~\ref{fig:problem_accuracy_comparison} and Figure~\ref{fig:lcbv6_performance_comparison}.

A closer analysis of the problems with notable improvement in Figure~\ref{fig:problem_accuracy_comparison} shows that RL training \textbf{does not introduce fundamentally new knowledge} beyond what is already present in the base model's outputs. This observation applies to both less challenging problems (where the base model already performs well) and more challenging ones. Instead, the improvements \textbf{mainly result from enhanced reasoning capabilities}. We identify three main areas of improvement:
\begin{itemize}
    \item \textbf{Enhanced Structural Organization}: Responses demonstrate a clearer logical flow and improved structural coherence.
    \item \textbf{Increased Attention to Details}: Models are more careful with edge cases and boundary conditions. This effect is especially clear in the Code-RL model, likely because boundary handling is important in programming tasks.
    \item \textbf{Improved Contextual Consistency}: RL-trained models are more accurate at integrating and summarizing previous reasoning steps. In contrast, the base model sometimes produces final answers based on incorrect intermediate reasoning even if some steps are correct, which leads to inconsistencies.
\end{itemize}

These findings further support our main claim: the main way RL improves model capability \textbf{is not by changing stored knowledge or basic skills} (such as arithmetic), \textbf{but by better integrating and optimizing existing abilities} through structured logical behavior such as reflection and planning. At the same time, this provides empirical support for the effectiveness of our proposed dual-token constraint training strategy.

\section{Licenses}
\label{app:licenses}

The licenses of used assets in this paper are listed as follows:
\begin{itemize}
    \item MATH: MIT License
    \item AMC23: Apache-2.0 License
    \item Minerva Math: No License
    \item OlympiadBench: MIT License
    \item verl: Apache-2.0 License
\end{itemize}


\end{document}